\documentclass[sigconf]{acmart}

\usepackage{xcolor}
\usepackage{algorithm}
\usepackage{booktabs, multirow} 
\usepackage{multirow}
\usepackage{algorithm}
\usepackage{amsthm}
\usepackage{amsmath}
\usepackage{mathtools}
\usepackage{bm}
\usepackage{tabularx}
\usepackage[noend]{algpseudocode}

\usepackage[utf8]{inputenc}
\usepackage{url}
\usepackage{booktabs}
\usepackage{bbding}
\usepackage{pifont}
\usepackage{wasysym}
\usepackage{utfsym}
\usepackage{fontawesome}

\usepackage{pdfpages}
\usepackage{graphicx}
\usepackage{caption}
\usepackage{subcaption}

\renewcommand\footnotetextcopyrightpermission[1]{} 

\acmDOI{} 
\acmISBN{} 

\begin{document}

\title[Abnormality Forecasting: Time Series Anomaly Prediction via Future Context Modeling]{Abnormality Forecasting: Time Series Anomaly Prediction via \\Future Context Modeling}



\author{Sinong Zhao}
\email{zhaosn@nbjl.nankai.edu.cn}
\affiliation{%
  \institution{College of Computer Science, TMCC, SysNet, DISSec, GTIISC, Nankai University}
  \country{China}
}

\author{Wenrui Wang}
\email{wangwenrui@nbjl.nankai.edu.cn}
\affiliation{%
\institution{College of Computer Science, TMCC, SysNet, DISSec, GTIISC, Nankai University}
  \country{China}
}

\author{Zhaoyang Yu}
\email{yuzz@nbjl.nankai.edu.cn}
\affiliation{%
\institution{College of Computer Science, TMCC, SysNet, DISSec, GTIISC, Nankai University}
  \country{China}
}

\author{Hongzuo Xu}
\email{hongzuoxu@126.com}
\affiliation{%
 \institution{Intelligent Game and Decision Lab (IGDL)}
 \country{China}
 }

\author{Qingsong Wen}
\email{qingsongedu@gmail.com}
\affiliation{%
  \institution{Squirrel AI}
  \country{China}
  }

\author{Gang Wang}
\email{wgzwp@nbjl.nankai.edu.cn}
\affiliation{%
\institution{College of Computer Science, TMCC, SysNet, DISSec, GTIISC, Nankai University}
  \country{China}
}

\author{Xiaoguang Liu}
\email{liuxg@nbjl.nankai.edu.cn}
\authornote{Corresponding authors.}
\affiliation{%
\institution{College of Computer Science, TMCC, SysNet, DISSec, GTIISC, Nankai University}
  \country{China}
}

\author{Guansong Pang}
\email{gspang@smu.edu.sg}
\authornotemark[1]
\affiliation{%
  \institution{School of Computing and Information Systems, Singapore Management University}
  \country{Singapore}
  }


\begin{abstract}
Identifying anomalies from time series data plays an important role in various fields such as infrastructure security, intelligent operation and maintenance, and space exploration.
Current research focuses on detecting the anomalies after they occur, which can lead to significant financial/reputation loss or infrastructure damage.
In this work we instead study a more practical yet very challenging problem, time series anomaly prediction, aiming at providing early warnings for abnormal events before their occurrence. 
To tackle this problem, we introduce a novel principled approach, namely \underline{f}uture \underline{c}ontext \underline{m}odeling (\textbf{FCM}). Its key insight is that \textit{the future abnormal events in a target window can be accurately predicted if their preceding observation window exhibits any \textbf{subtle} difference to normal data}. To effectively capture such differences, FCM first leverages long-term forecasting models to generate a discriminative future context based on the observation data, aiming to amplify those subtle but unusual difference. It then models a normality correlation of the observation data with the forecasting future context to complement the normality modeling of the observation data in foreseeing possible abnormality in the target window. A joint variate-time attention learning is also introduced in FCM to leverage both temporal signals and features of the time series data for more discriminative normality modeling in the aforementioned two views.
Comprehensive experiments on five datasets demonstrate that FCM gains good recall rate (70\%+) on multiple datasets and significantly outperforms all baselines in $F_{1}$ score. 
Code is available at https://github.com/mala-lab/FCM.

\end{abstract}




\keywords{Time Series, Anomaly Prediction, Anomaly Detection, Forecasting}



\maketitle

\section{Introduction}

\begin{figure}[!t]
\centering
\centerline{\includegraphics[height=6.5cm]{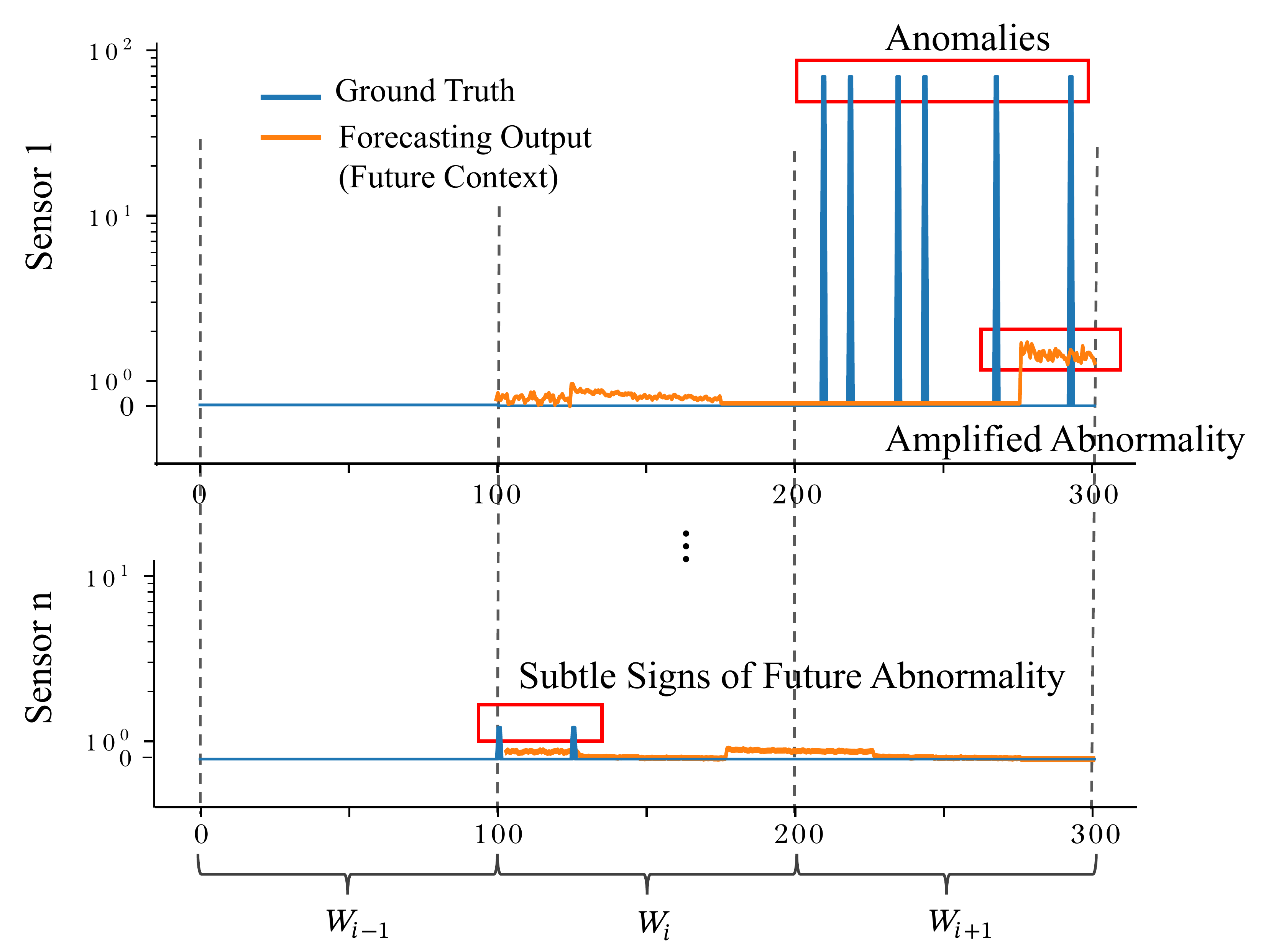}}
\caption{Illustration of the key insight of FCM. For effective anomaly prediction, FCM assumes that there are subtle signs (in the bottom time series) at the observation window for future abnormal events in a target window. These signs are typically too subtle to be detected by TSAD models. FCM aims to leverage long-term time series forecasting models to amplify these signs and associate them as a future context with the observation temporal signals for accurate anomaly prediction. 
}
\label{fig:intro}
\end{figure}

Identifying anomalies from time series data is highly demanded in practice~\cite{blazquez2021review,gupta2013outlier,jin2023survey}. Detecting anomalies that have already occurred is valuable, but the occurrence of the abnormal events can lead to significant financial/reputation loss or infrastructure damage. Accurately predicting these anomalies in advance, i.e., having early warnings of the anomalies, can effectively mitigate such adverse effects.
Time series anomaly prediction holds significant importance across various fields.
For instance, in the monitoring of critical infrastructure systems, anomaly prediction ensures the safety and stability of water treatment systems by preventing abnormal events such as water supply/pollution accidents~\cite{mathur2016swat}; in machine operation and maintenance~\cite{xiao2018disk,xie2019dfpe}, it provides early warnings for potential failures in servers, hard drives, and other equipment, enabling preemptive measures to avoid financial/reputation losses due to the failures; in space exploration, it alerts potential operational/logistic issues beforehand, providing time for handling the issues early, thereby guaranteeing smooth progression of exploration missions~\cite{hundman2018detecting}. Thus, this work focuses on addressing the anomaly prediction problem.

Numerous methods have been introduced for time series anomaly detection (TSAD), such as reconstruction-based methods~\cite{jie2024disentangled,li2023prototype,song2024memto,lai2024nominality,xu2021anomaly,su2019robust,li2021multivariate}, contrastive learning-based methods~\cite{yang2023dcdetector,kim2023contrastive}, one-class classification methods~\cite{manevitz2001one,xu2024calibrated}, and graph neural network-based methods~\cite{deng2021graph,chen2022graphad,zhang2022grelen}. 
TSAD focuses on detecting the abnormal events after they occur, so they assume that clear abnormal patterns exist in the observational time series data, rendering them ineffective for anomaly prediction. This is because the signs of having abnormal events in the future time points are typically subtle, such as the data in $W_{i}$ in Fig. \ref{fig:intro}. 

To tackle this challenge, we introduce a novel principled approach, namely \underline{f}uture \underline{c}ontext \underline{m}odeling (\textbf{FCM}). Its key insight is that \textit{the future abnormal events in a target window can be accurately predicted if their preceding observation window exhibits any \textbf{subtle} difference to normal data}.  To effectively capture such differences, FCM first leverages long-term forecasting (LTTSF) models~\cite{nie2022time,wangcard2024,liu2023itransformer,wu2022timesnet} to generate a discriminative future context based on the current time window data, aiming to amplify those subtle but unusual difference. This is because i) the LTTSF techniques can often accurately predict normal time points, resulting in smooth, accurate forecasting time points, but ii) they are difficult to accurately forecast the abnormal time points, leading to fluctuated, exceptional time points. For example, as shown in Fig. \ref{fig:intro}, at the normal time points $W_i$, the forecasting output given $W_{i-1}$ does not exhibit any abnormality, but it starts behaving abnormally at $W_{i+1}$ when observing $W_i$ where subtle sign of anomalies is presented. The unusual forecasting signals at $W_{i+1}$ on the top help largely amplify the subtle abnormality signs at current window $W_i$.

These forecasting time points are treated as future contexts, which are then utilized by FCM to model its normality correlation with the observation data for foreseeing possible abnormality in the target window. That is, FCM takes both the current observation time points and its forecasting time points as joint input, upon which a joint data reconstruction model is built. Due to the subtle anomaly signs in the current time points and their amplification via the forecasting, the joint reconstruction error is expected to be large if the future target window contains abnormal time points, and it would be small otherwise. Additionally, FCM also adds a module that individually model the current time points so as to capture any sign of future abnormality without being affected by the forecasting time points. Thus, FCM foresees future abnormality through two different yet complementary views: one view is on modeling any subtle abnormality sign at the current time points while another view is based on the normality correlation between the current and forecasting time points.

Furthermore, to support a more discriminative reconstruction for future normal and abnormal time points, we propose a multi-dimensional self-attention module to obtain correlations from both the feature dimension and the temporal dimension. The subtle anomaly signs often have difficulty establishing correlations with the future abnormal time points in at least one of these two dimensions, while the normal time points are strongly correlated at both the feature and temporal dimensions. Thus, the multi-dimensional self-attention approach helps increase the difficulty in the joint data reconstruction when the future time window contains abnormal time points. By contrast, it enables better data reconstruction when the future time points are all normal.

Our contributions are summarized as follows:

\begin{itemize}
\item We explore an important yet under-explored problem, time series anomaly prediction, aiming to promote a more practical setting for anomaly identification in time series data.
\item 
We then propose the novel anomaly prediction approach FCM that aims to learn and leverage future context to amplify subtle abnormality signs at the observation window and provide more discriminative features for anomaly prediction. To our best knowledge, this is the first work that models the normality correction between the observation and forecasting time points for anomaly prediction.

\item 
We further introduce a variate-time multi-dimensional self-attention module, in which the correlation of time series data can be effectively modeled from both of the feature and temporal dimensions, enabling a more discriminative joint modeling of the current and forecasting time points in FCM.

\item 
We establish evaluation protocols using five widely-used TSAD datasets and perform comprehensive experiments to compare our FCM with 16 anomaly prediction methods. FCM gains good recall rate (70\%+) on multiple datasets and consitently outperforms the competing methods in $F_{1}$ score. 

\end{itemize}

\section{Related Work}
\subsection{Time Series Anomaly Detection}

Studies on identifying anomalies in time series data are focused on the task of detection rather than prediction. Numerous methods have been introduced in this line of research, which can be roughly categorized into the following five groups.
1) Classical Methods: 
   Classical methods are not specific to time series data but are generally applicable to all data types, such as OCSVM~\cite{manevitz2001one}, iForest~\cite{liu2008isolation,xu2023deep}, and DAGMM~\cite{zong2018deep}.
2) Reconstruction-based Methods: 
   The main idea of reconstruction-based methods is to learn the manifold of normal classes. When encoding and reconstructing data using autoencoders or other methods, anomalies cannot be properly reconstructed due to their significant differences from the normal manifold. Many reconstruction architectures have been introduced \cite{su2019robust,li2021multivariate,xu2021anomaly,jie2024disentangled,li2023prototype,song2024memto}. For example, 
   Anomaly Transformer (AnomTrans)~\cite{xu2021anomaly} targets the correlation of consecutive times with a transformer network, discovering weak relationships between anomalies and the entire sequence. 
   Methods such as DA-VAE~\cite{jie2024disentangled}, 
   PUAD~\cite{li2023prototype}, and MEMTO~\cite{song2024memto} learn diverse normal patterns via prototype- or memory-augmented reconstruction.
3) Contrastive Learning-based Methods: These methods 
    learn temporal normal patterns via contrastive learning, such as DCdetector~\cite{yang2023dcdetector} and CTAD~\cite{kim2023contrastive}. 
4) Graph Neural Network-based Methods: 
   They learn the structure of existing relationships between variables using graph neural networks and distinguish anomalies by predicting the future values of each sensor, such as GDN~\cite{deng2021graph}.
5) Generative Adversarial Network-based Methods: The methods in this group focus on utilizing adversarial training or adversarially generated time series data to train the detection models, such as 
   USAD~\cite{audibert2020usad} 
   and BeatGAN~\cite{zhou2019beatgan}.

\subsection{Time Series Forecasting}
Long-term time series forecasting (LTTSF) is a classic task in time series analysis. 
It involves extracting the core patterns embedded in extensive data and estimating changes over a long period in the future. 
In recent years, numerous studies have attempted to apply transformer models to LTTSF~\cite{wen2023transformers}. 
For example, 
   Informer~\cite{zhou2021informer} aims to adopt distillation techniques together with self-attention to effectively extract the most crucial time points for the forecasting.
   Autoformer~\cite{wu2021autoformer} draws on ideas from traditional time series analysis methods and incorporates decomposition and auto-correlation into the network.
   FEDformer~\cite{zhou2022fedformer} uses a Fourier-enhanced structure to achieve linear complexity.
   PatchTST~\cite{nie2022time} treats patches as input units, preserving the semantics of each block in the time series data, thereby utilizing the transformer structure more effectively to achieve good results. Subsequent works have mostly followed the patch concept.
   iTransformer~\cite{liu2023itransformer} embeds each time series as variable tokens, employing the attention mechanism to handle multivariate correlations and using a feed-forward network for sequence encoding. It adopts a reversed perspective on time series, embedding the entire time series of each variable independently into tokens, thereby expanding the local receptive field.
These forecasting models may be adapted for anomaly prediction via a simple reconstruction module, but they lack the designs that focus on learning the prevalent patterns from the data, leading to ineffective anomaly prediction. Limited work has been done on anomaly prediction. Jhin et al.~\cite{jhin2023precursor} explore the detection of early signs of abnormality in time series data to provide a unified framework for anomaly detection and prediction, but its future prediction is restricted to very short time period, leading to a task similar to anomaly detection.

\begin{figure}[!t]
\centering
\centerline{\includegraphics[height=3.4cm]{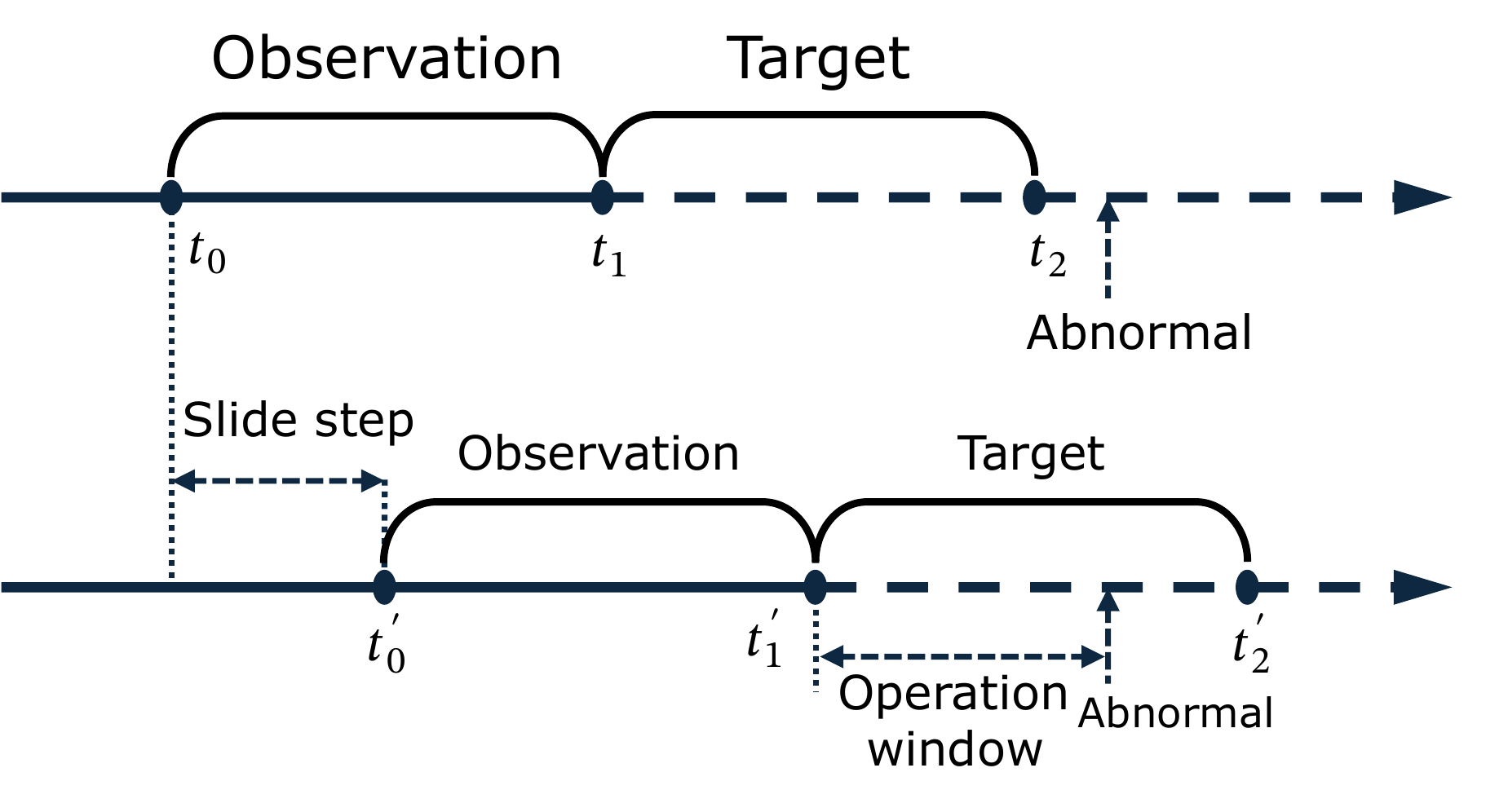}}
\caption{The studied setting. Given an observational time window from $t_i$ to $t_{i+1}$, we aim to predict whether there would be abnormal events in the target window from $t_{i+1}$ to $t_{i+2}$. The two windows are slid with a fixed step size (see Sec. \ref{subsec:setting}). }
\label{fig:setting}
\end{figure}

\section{Future Context Modeling}
\subsection{Problem Statement}\label{subsec:setting}

Let $x_{1:T} = \{x_1, \ldots, x_t, \ldots, x_T\}\in \mathbb{R}^{D\times T} $ be a multivariate time series, 
where $x_t\in \mathbb{R}^D$ is a multi-dimensional vector, $D$ represents the dimensionality of the multivariate time series, and $T$ is the length of the time series.

We first formalize the anomaly prediction task for multivariate time series data. 
As shown in Fig.~\ref{fig:setting}, a sliding window is leveraged.
Let \textit{observation window} be $x_i = [x_{t_{i1}}, \ldots, x_{t_{iL}}]$, where $t_{ij} = i \times S + j - 1 $.
The window size is \(L\) and the sliding step is \(S\).
The observation window serves as the input to our neural network.
The next non-overlapping window $[x_{t_{i(L+1)}}, \ldots, x_{t_{i(2L)}}]$ is the target window.
Given an input observation window \( x_i \), our network determines whether each time point in the target window is an anomaly.
There are a total of \( \lfloor (T-L)/S \rfloor+1 \) observation windows.
The period before an anomaly occurs in the target window is called \textit{operation window}, in which we take actions to mitigate the impact if the prediction results have anomalies.
The length of the actionable time is adjusted by the sliding step size $S$. 
Note that the step size needs to be smaller than the window size to ensure sufficient actionable time. 

\begin{figure*}[!t]
\centering
\centerline{\includegraphics[height=8.6cm]{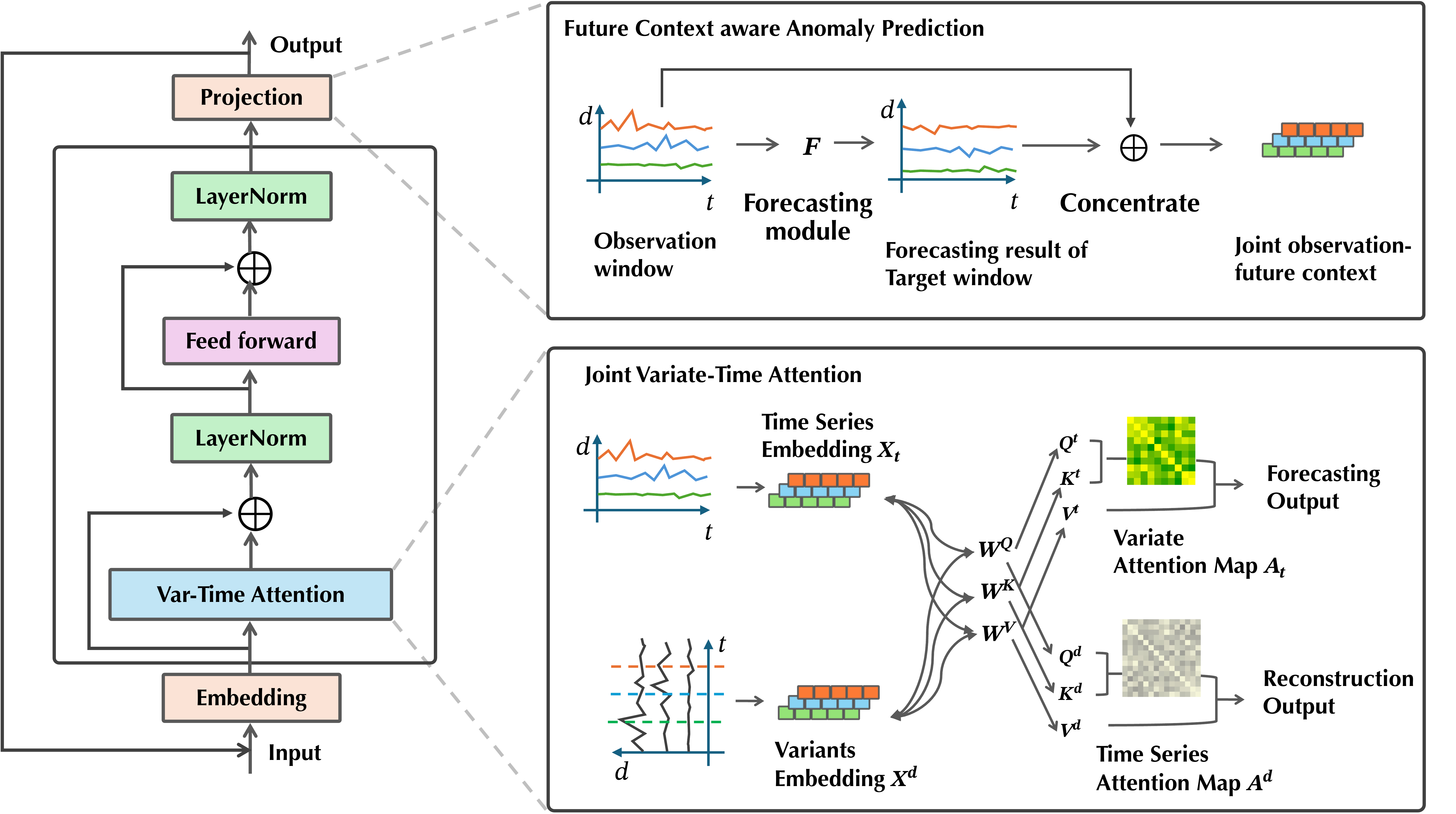}}
\caption{Overview of FCM.
The future context-aware anomaly prediction in FCM leverages the output of the forecasting task to enhance the representation of the current window, generating a new discriminative representation of the observation data for joint normality modeling.
The var-time attention module learns the relationships between temporal signals and feature values of time series data to enable more discriminative normality learning.
}
\label{fig:flow}
\end{figure*}

\subsection{Overview of FCM}
This paper proposes a novel time-series anomaly prediction method FCM. To achieve accurate abnormal event prediction, it leverages long-term forecasting models to generate a discriminative future context for modeling its normality correlation with the data in the observation window.
As shown in Fig.~\ref{fig:flow}, FCM contains two important components: a future context-aware anomaly prediction module and a joint variate-time (var-time) attention module.
The anomaly prediction module consists of normality modeling from two views. The first view is to model the normality association of observation window data with a forecasting future context to achieve accurate abnormal prediction through joint data reconstruction. Another view is on the normality modeling of the observation data to capture any abnormality signs without being affected by the forecasting output.
The var-time attention module constrains anomalies by modeling from the temporal signals and feature dimensions of time series data, increasing the reconstruction difference between normal and abnormal.
Below we introduce the details of each module.

\subsection{Future Context-aware Anomaly Prediction}
\label{sec:fcm}

FCM aims to forecast a future context for more accurate anomaly prediction. The key intuition behind is that time series forecasting results can behave very differently, depending on whether there are abnormal time points or not. This allows the amplification of abnormality signs at the observation window. Thus, we design a LTTSF module $F$ in FCM. 
This module predicts the target window by learning the continuous changes in two adjacent windows in the training data with normal data. 
For current window $x_i$, the forecasting results for the target window through $F$ can be defined as $\tilde{x}_{t_{i(L+1)}:t_{i(2L)}} = f({x}_{t_{i1}:t_{iL}};\Theta_{fore})$, with the parameters $\Theta_{fore}$ optimized using a mean squared error (MSE) loss:
\begin{equation}
   \mathcal{L}_{fore} = \left \|  \tilde{x}_{t_{i(L+1)}:t_{i(2L)}}-{x}_{t_{i(L+1)}:t_{i(2L)}}\right \|^2_2.
   \label{eq:foreloss}
\end{equation}

As shown in Fig.~\ref{fig:intro}, although the forecasting module has significant errors in predicting the abnormal signals, there are discriminative forecasting outputs for normal and abnormal time points in the target windows, i.e., smooth, accurate forecasting for normal data signals in the target window $W_i$ vs. fluctuated, exceptional forecasting results for abnormal data signals in the target window $W_{i+1}$. Therefore, the forecasting result $\tilde{x}_{t_{i(L+1)}:t_{i(2L)}}$ can be seen as an amplifier that leverages the subtle abnormality signals in the observation window $W_i$ to generate a discriminative future context at $W_{i+1}$.
FCM then models the normality correlation between the observation window data and its forecasting future context via a linear layer $g$ parameterized by $\Theta_g$ through joint observation-future context signal reconstruction:
\begin{gather}
   \label{eq:interloss}
   C_{i} = g([x_{t_{i(1)}:t_{i(L)}} \cdot \tilde{x}_{t_{i(L+1)}:t_{i(2L)}}]; \Theta_g),\\
   \mathcal{L}_{c} = \left \|  \hat{C}_{i}-C_{i}\right \|^2_2.
\end{gather}
where $[\cdot]$ means a concatenation of the two inputs and $\hat{C}=h(C;\Theta_c)$ refers to the reconstruction of $C$ by a decoder $h$ parameterized by $\Theta_c$. 
Our LTTSF module can predict accurately for most normal points. Thus, the observation-future context data consisting of $x_{t_{i(1)}:t_{i(L)}}$ and $\tilde{x}_{t_{i(L+1)}:t_{i(2L)}}$ are strongly correlated, if $x_{t_{i(1)}:t_{i(L)}}$ does not contain future abnormality signs. This correlation is broken otherwise, since the forecasting output $\tilde{x}_{t_{i(L+1)}:t_{i(2L)}}$ would not follow specific distribution patterns. As a result, the reconstruction for normal events in the target window would be small, and it would be large for future abnormal events. 


Additionally, we also learn the normality pattern from the observation window individually to model the original sign anomalous signals. This is to complement the normality modeling from the observation-future context view. It is done by a usual reconstruction of the observation data $x_i$:
\begin{equation}
   \mathcal{L}_{det} = \left \|  \hat{x}_{t_{i(1)}:t_{i(L)}}-x_{t_{i(1)}:t_{i(L)}}\right \|^2_2,
   \label{eq:adloss}
\end{equation}
where $\hat{x}_{t_{i(1)}:t_{i(L)}}$ refers to the reconstruction result of $x_{t_{i(1)}:t_{i(L)}}$.


\subsection{Joint Variate-Time Attention Mechanism}
In Sec.~\ref{sec:fcm}, we use data reconstruction to model the normality patterns from two different views.
However, traditional reconstruction methods may also reconstruct subtle anomaly signals in both views. 
To increase the difficulty of the data reconstruction that involves abnormal signals, we propose a joint variate-time (var-time) attention mechanism. 
This mechanism learns data correlations from both the temporal and feature dimensions, making it difficult to establish connections for current windows that exhibit anomalies in at least one dimension, thereby enabling more accurate anomaly prediction.


First, we set the self-attention from the variate dimension. 
Following the early work iTransformer~\cite{liu2023itransformer}, 
time series of each sensor is embedded into a $d_{model}$ dimension space as a token, in which the original data $X^{D\times T}$ can be defined as $X_t = \{x_t^0,x_t^1,...,x_t^D\}\in R^{D\times d_{model}}$.
We then feed the embedded information into the encoder of the transformer. Each head of the multi-head attention utilizes $\mathbf{W}^Q, \mathbf{W}^K, \mathbf{W}^V$ to transform $\mathbf{X}_t$ into $\mathbf{Q}_t, \mathbf{K}_t, \mathbf{V}_t \in \mathbb{R}^{D \times h \times d_k}$ by matrix multiplication, with 
\begin{equation}
   \mathbf{Q}_t = \mathbf{W}^Q\mathbf{X}_t, \mathbf{K}_t = \mathbf{W}^K\mathbf{X}_t, \mathbf{V}_t = \mathbf{W}^V\mathbf{X}_t,
\end{equation}
where $\mathbf{W}^Q, \mathbf{W}^K\in \mathbb{R}^{d_{model} \times (h \times d_k)}, \mathbf{W}^V\in \mathbb{R}^{d_{model} \times (h \times d_v)}$,
$h$ is the number of multi-heads, and $d_k$ is the dimension of each head. For each head, the dot product of the query and the key is calculated, and then the attention map for the variate dimension is obtained via:
\begin{equation}
   \mathbf{A}_t = softmax(\frac{\mathbf{Q}_t \mathbf{K}_t^T}{\sqrt{d_k}}),
\end{equation}
where $\mathbf{A}_t\in \mathbb{R}^{h \times D \times D}$.
We then obtain the outputs via a weighted sum of the values $\mathbf{V}_t$ and the attention weights $\mathbf{A}_t$:
\begin{equation}
   \mathbf{O}_t = \mathbf{A}_t \mathbf{V}_t\in \mathbb{R}^{D \times d_{model}},
\end{equation}
which is used to enable our time series forecasting in Sec. \ref{sec:fcm}.

Correspondingly, we feed the original data $X^{D\times T}$ to realize anomaly prediction in the temporal dimension.
It is defined as $X^d = \{x_0^d,x_1^d,...,x_T^d\}\in R^{d_{model}\times T}$ after values of all sensors at the same moment are embedded into $d_{model}$-dimension space. We establish the anomaly prediction module based on the correlation between time points. 
Since the anomaly detection and the prediction modules share one encoder, we can obtain $\mathbf{Q}^d, \mathbf{K}^d, \mathbf{V}^d \in \mathbb{R}^{L \times h \times d_k}$, which is the same as in the forecasting module:
\begin{gather}
   \mathbf{Q}^d = \mathbf{W}^Q\mathbf{X}^d, \mathbf{K}^d = \mathbf{W}^K\mathbf{X}^d, \mathbf{V}^d = \mathbf{W}^V\mathbf{X}^d,\\
   \mathbf{A}^d = softmax(\frac{\mathbf{Q}^d (\mathbf{K^d})^T}{\sqrt{d_k}})\in \mathbb{R}^{h \times L \times L}.
\end{gather}
$\mathbf{O}^d = \mathbf{A}^d \mathbf{V}^d\in \mathbb{R}^{L \times d_{model}},$ which we use it to build the reconstruction modules in in Sec. \ref{sec:fcm}.

This self-attention mechanism helps learn the correlations of the temporal and feature dimensions at the same time, so $\mathbf{O}^d$ contains both temporal and variate corresponding, make it more discriminative of the reconstruction errors between the normal and subtle abnormal in both reconstruction modules of our future context-aware anomaly prediction.

\begin{table*}[htbp]
  \centering
  \caption{
  Precision $P$, Recall $R$, and $F_1$ score of anomaly prediction results on five multivariate real-world datasets. 
  All results are represented in percentages (\%). The best performance for each metric on each dataset is highlighted in bold, and the runner-up $F_1$ score are underlined. 
  }
  \scalebox{0.8}{

    \begin{tabular}{l|rrr|rrr|rrr|rrr|rrr}
    \toprule
    \toprule
    \multicolumn{1}{c|}{} & \multicolumn{3}{c|}{SMD} & \multicolumn{3}{c|}{MSL} & \multicolumn{3}{c|}{SMAP} & \multicolumn{3}{c|}{SWaT} & \multicolumn{3}{c}{PSM} \\
    \midrule
    \multicolumn{1}{l|}{Method} & \multicolumn{1}{c}{P} & \multicolumn{1}{c}{R} & \multicolumn{1}{c|}{$F_1$} & \multicolumn{1}{c}{P} & \multicolumn{1}{c}{R} & \multicolumn{1}{c|}{$F_1$} & \multicolumn{1}{c}{P} & \multicolumn{1}{c}{R} & \multicolumn{1}{c|}{$F_1$} & \multicolumn{1}{c}{P} & \multicolumn{1}{c}{R} & \multicolumn{1}{c|}{$F_1$} & \multicolumn{1}{c}{P} & \multicolumn{1}{c}{R} & \multicolumn{1}{c}{$F_1$} \\
    \midrule
    OCSVM & 7.06  & 11.23  & 8.67  & 39.21  & 46.15  & 42.40  & 23.41  & 34.16  & 27.78  & 6.72  & 17.92  & 9.77  & 3.51  & 26.30  & 6.19  \\
    iForest & 4.52  & 3.27  & 3.80  & 38.94  & 20.34  & 26.72  & 29.28  & 46.29  & 35.87  & 7.34  & 10.19  & 8.53  & 17.34  & 23.17  & 19.84  \\
    BeatGAN & 27.36  & 38.35  & \underline{31.94} & 58.36  & 58.99  & \underline{58.67} & 34.49  & 56.37  & \underline{42.80} & 15.05  & 39.67  & 21.82  & 42.30  & 36.54  & \underline{39.21 } \\
    DAGMM & 2.12  & 43.10  & 4.05  & 47.30  & 53.24  & 50.10  & 20.78  & 59.13  & 30.75  & 0.26  & 7.78  & 0.50  & 5.66  & 45.18  & 10.06  \\
    OmniAnomaly & 0.08  & 0.17  & 0.11  & 10.09  & 10.07  & 10.08  & 2.55  & 5.83  & 3.55  & 0.37  & 14.02  & 0.72  & 15.77  & 13.00  & 14.25  \\
    THOC  & 2.49  & 6.06  & 3.53  & 23.55  & 35.27  & 28.24  & 21.39  & 30.56  & 25.17  & 0.61  & 24.79  & 1.18  & 2.60  & 12.23  & 4.29  \\
    InterFusion & 1.58  & 3.43  & 2.16  & 23.65  & 28.38  & 25.80  & 19.71  & 54.09  & 28.89  & 0.60  & 15.51  & 1.16  & 4.85  & 3.36  & 3.97  \\
    LSTM-ED & 6.72  & 11.05  & 8.35  & 37.12  & 42.36  & 39.57  & 14.37  & 39.12  & 21.02  & 0.83  & 25.36  & 1.60  & 5.07  & 12.00  & 7.13  \\
    MSCRED & 1.53  & 17.03  & 2.81  & 6.26  & 24.05  & 9.94  & 3.65  & 41.46  & 6.71  & 0.56  & 6.15  & 1.02  & 2.85  & 22.02  & 5.05  \\
    GDN   & 4.31  & 8.16  & 5.64  & 24.85  & 38.51  & 30.21  & 17.06  & 45.53  & 24.82  & 3.08  & 21.64  & 21.64  & 1.97  & 6.35  & 3.01  \\
    COUTA & 3.62  & 35.93  & 6.58  & 36.37  & 35.61  & 35.98  & 14.56  & 37.86  & 21.03  & 0.14  & 3.38  & 0.27  & 2.49  & 19.34  & 4.41  \\
    FEDformer & 3.08  & 2.84  & 2.96  & 43.52  & 24.22  & 31.12  & 15.87  & 21.25  & 18.17  & 4.08  & 21.16  & 6.85  & 5.24  & 12.83  & 7.44  \\
    Autoformer & 3.08  & 2.84  & 2.96  & 43.01  & 23.68  & 30.55  & 14.02  & 19.13  & 16.18  & 4.07  & 21.16  & 6.82  & 5.18  & 12.83  & 7.38  \\
    TimesNet & 14.27  & 12.51  & 13.33  & 39.00  & 24.22  & 29.89  & 19.76  & 24.24  & 21.77  & 12.97  & 26.07  & 17.32  & 30.21  & 10.80  & 15.91  \\
    AnomTrans & 25.19  & 34.46  & 29.10  & 39.71  & 57.86  & 47.10  & 30.51  & 51.51  & 38.32  & 24.01  & 61.03  & \underline{34.46} & 41.55  & 36.45  & 38.84  \\
    DCdetector & 17.14  & 21.53  & 19.08  & 21.17  & 11.50  & 14.91  & 6.82  & 8.44  & 7.54  & 5.49  & 13.26  & 7.77  & 7.91  & 8.02  & 7.96  \\
    \midrule
    FCM (Ours)  & \textbf{34.23} & \textbf{54.13} & \textbf{41.93} & \textbf{65.50} & \textbf{82.90} & \textbf{73.18} & \textbf{39.46} & \textbf{76.09} & \textbf{51.97} & \textbf{25.62} & \textbf{74.97} & \textbf{38.19} & \textbf{56.07} & \textbf{65.92} & \textbf{60.60 } \\
    \bottomrule
    \bottomrule
    \end{tabular}%

    }
  \label{tab:result}%
\end{table*}%

\begin{table}[htbp]
  \centering
  \caption{FCM vs. the two best-performing contenders in more evaluation metrics. The results of other methods are available in Table~\ref{tab:all metrics} in the appendix.
}
\scalebox{0.68}{

    \begin{tabular}{c|l|cccccccc}
    \toprule
    \toprule
    \multicolumn{1}{l|}{Dataset} & Method  & \multicolumn{1}{c}{Acc} & \multicolumn{1}{c}{$F_1$} & \multicolumn{1}{c}{Aff-P} & \multicolumn{1}{c}{Aff-R} & \multicolumn{1}{c}{R\_A\_R} & \multicolumn{1}{c}{R\_A\_P} & \multicolumn{1}{c}{V\_ROC} & \multicolumn{1}{c}{V\_PR} \\
    \midrule
    \multirow{3}[2]{*}{SMD} & BeatGAN & 87.31  & 31.94  & 50.06  & \textbf{97.56} & \textbf{64.28} & 25.88  & \textbf{63.67} & 25.29  \\
          & AnomTrans & 98.41  & 29.10  & 49.39  & 90.09  & 57.04  & 22.07  & 56.12  & 21.15  \\
          & FCM (Ours)  & \textbf{98.58} & \textbf{41.93} & \textbf{50.46} & 93.24  & 61.37  & \textbf{30.79} & 59.96  & \textbf{29.33 } \\
    \midrule
    \multirow{3}[2]{*}{MSL} & BeatGAN & 82.90  & 58.67  & 49.81  & \textbf{98.56} & 72.44  & 38.90  & 73.69  & 40.22  \\
          & AnomTrans & 98.04  & 47.10  & 50.34  & 93.84  & 66.27  & 46.88  & 65.09  & 45.76  \\
          & FCM (Ours)  & \textbf{98.66} & \textbf{73.18} & \textbf{50.51} & 96.12  & \textbf{75.34} & \textbf{59.81} & \textbf{74.63} & \textbf{59.14 } \\
    \midrule
    \multirow{3}[2]{*}{SMAP} & BeatGAN & 92.17  & 42.80  & 50.09  & \textbf{99.43} & \textbf{74.17} & 33.81  & \textbf{74.86} & 34.52  \\
          & AnomTrans & 98.50  & 38.32  & \textbf{50.54} & 97.97  & 66.40  & 33.02  & 65.70  & 32.32  \\
          & FCM (Ours)  & \textbf{98.81} & \textbf{51.97} & 49.95  & 97.95  & 73.97  & \textbf{44.77} & 72.63  & \textbf{43.45 } \\
    \midrule
    \multirow{3}[2]{*}{SWaT} & BeatGAN & 82.84  & 21.82  & 49.77  & \textbf{99.23} & 73.60  & 33.69  & \textbf{74.06} & 34.15  \\
          & AnomTrans & \textbf{98.92} & 34.46  & 50.27  & 87.15  & 71.14  & 34.85  & 71.66  & 35.39  \\
          &  FCM (Ours)  & 98.91  & \textbf{38.19} & \textbf{50.64} & 95.35  & \textbf{74.11} & \textbf{37.71} & 73.03  & \textbf{36.64 } \\
    \midrule
    \multirow{3}[2]{*}{PSM} & BeatGAN & 91.82  & 39.21  & 50.16  & \textbf{93.90} & \textbf{66.97} & 33.46  & \textbf{65.97} & 32.36  \\
          & AnomTrans & 97.87  & 38.84  & \textbf{51.54} & 78.03  & 58.54  & 32.71  & 57.70  & 31.94  \\
          &  FCM (Ours)  & \textbf{98.40} & \textbf{60.50} & 49.86  & 80.15  & 65.93  & \textbf{46.82} & 64.24  & \textbf{45.06 } \\
    \bottomrule
    \end{tabular}%

}
  \label{tab:vus}%
\end{table}%

\begin{table}[htbp]
  \centering
  \caption{$F_1$ score performance for the anomaly detection task. 
  All results are in \%, and the best ones are in Bold, and the runner-up performer are underlined.}
  \scalebox{0.78}{
    \begin{tabular}{l|ccccc}
    \toprule
    \toprule
    \multicolumn{1}{c|}{Dataset} & SMD   & MSL   & SMAP  & SWaT  & PSM \\
    \midrule
    OCSVM & 63.27  & 82.66  & 61.75  & 80.70  & 88.35  \\
    iForest & 51.01  & 74.79  & 67.59  & 77.82  & 87.82  \\
    BeatGAN & \underline{78.37} & \underline{89.97} & 91.15  & 92.70  & 85.78  \\
    DAGMM & 31.24  & 86.98  & 67.03  & 74.87  & 85.11  \\
    OmniAnomaly & 21.56  & 57.68  & 71.69  & 56.62  & 78.10  \\
    THOC  & 26.95  & 52.72  & 42.62  & 56.70  & 73.83  \\
    InterFusion & 12.93  & 62.58  & 88.21  & 39.65  & 40.11  \\
    LSTM-ED & 34.28  & 57.56  & 54.39  & 57.04  & 77.71  \\
    MSCRED & 16.18  & 59.09  & 56.64  & 63.64  & 43.86  \\
    GDN   & 66.35  & 50.80  & 73.17  & 80.46  & 88.42  \\
    COUTA & 63.73  & 59.42  & 70.04  & 46.32  & 69.89  \\
    FEDformer & 44.98  & 67.74  & 59.93  & 84.16  & 87.39  \\
    Autoformer & 44.85  & 71.40  & 73.27  & 83.40  & 87.29  \\
    TimesNet & 76.27  & 80.75  & 69.77  & 91.91  & 92.79  \\
    AnomTrans & 75.29  & 86.16  & 92.11  & 93.17  & 94.83  \\
    DCdetector & 77.42  & 81.87  & \underline{93.10} & \underline{94.82} & \underline{95.67 } \\
    FCM (Ours)  & \textbf{81.36} & \textbf{92.01} & \textbf{95.11} & \textbf{95.55} & \textbf{96.81 } \\
    \bottomrule
    \bottomrule
    \end{tabular}%
    }
  \label{tab:detect}%
\end{table}%

\subsection{Anomaly Prediction Using FCM}
\noindent\textbf{Training}. 
We use a sequential updating strategy to provide feedback for the three losses involved in the task.
Specifically, 
we update the network using $\mathcal{L}_{det}$ and $\mathcal{L}_{fore}$ for a period of time before incorporating $\mathcal{L}_{c}$. 
This approach ensures that the forecasting module's results are relatively stable. 
If we incorporate $\mathcal{L}_{c}$ from the start, the large errors in time series forecasting results could adversely affect the forecasting results.

\noindent\textbf{Inference}.
Given the observation window data $x_{t_{i(1)}:t_{i(L)}}$, we obtain the forecasting result of the target window from our LTTSF module.
We then concatenate the forecasting future context with the observation data and embed it into a low-dimensional space.
We use the embedding $C_{i}$ as the representation of the observation data. 
The reconstruction error obtained from this representation is used as the anomaly score for anomaly prediction:
\begin{equation}
   Score(W_i) = \left \|  \hat{C}_{i}-C_{i}\right \|^2_2.
   \label{eq:score}
\end{equation}

\section{Experiments}

\subsection{Settings}
\subsubsection{Datasets}

Five widely-used TSAD datasets are utilized in the experiments~\cite{xu2021anomaly,yang2023dcdetector}. They are adapted to anomaly prediction task by shifting the ground truth of abnormal events to a proceeding window.

\begin{itemize}
    \item \textbf{Server Machine Dataset (SMD)} is a five-week dataset collected from a large Internet company, featuring 38 dimensions and resource utilization access traces from 28 machines~\cite{su2019robust}.

    \item \textbf{Mars Science Laboratory dataset (MSL)} is collected by NASA and contains sensor and actuator data from the Mars rover~\cite{hundman2018detecting}. It includes 55 dimensions and 27 entities, featuring telemetry anomaly data derived from event surprise anomaly (ISA) reports from the spacecraft monitoring system.

    \item \textbf{Soil Moisture Active Passive dataset (SMAP)} is also collected by NASA and includes soil samples and telemetry information from the Mars rovers~\cite{hundman2018detecting}. It has 25 dimensions and, compared to MSL, contains more point anomalies.

    \item \textbf{Safe Water Treatment dataset (SWaT)} is a 51-dimensional sensor-based dataset collected from continuously operating critical infrastructure systems~\cite{mathur2016swat}. It includes 11 consecutive days of data: 7 days of normal operation and 4 days of periodic attacks, totaling 41 attacks. The time granularity is 1 second.
    \item \textbf{PSM} (Pooled Server Metrics Dataset) is a public dataset from eBay server machines with 25 dimensions~\cite{abdulaal2021practical}.
\end{itemize}

\subsubsection{Competitors} 
We employ sixteen anomaly detectors as competing methods, including
the classic anomaly detection methods (OCSVM~\cite{manevitz2001one}, iForest~\cite{liu2008isolation}, DAGMM~\cite{zong2018deep}, and THOC~\cite{shen2020timeseries}), generative methods (BeatGAN~\cite{zhou2019beatgan} and COUTA~\cite{xu2024calibrated}), reconstruction-based methods (OmniAnomaly~\cite{su2019robust},  InterFusion~\cite{li2021multivariate}, LSTM-ED~\cite{malhotra2016lstm}, MSCRED~\cite{zhang2019deep}, AnomTrans~\cite{xu2021anomaly}, and DCdetector~\cite{yang2023dcdetector}), graph neural network-based methods (GDN~\cite{deng2021graph}), and time series analysis-based models (FEDformer~\cite{zhou2022fedformer}, Autoformer~\cite{wu2021autoformer}, and TimesNet~\cite{wu2022timesnet}).
Among these methods, thirteen anomaly detection models can be directly applied to anomaly prediction by modeling normality on the observation window data. Additionally, the remaining three time series analysis-based models are modified and adapted for anomaly prediction via a data reconstruction module on the forecasting data.

\subsubsection{Evaluation Metrics}\label{sec:metrics}

Time-series anomaly prediction presents distinct challenges compared to the well-established anomaly detection task, necessitating a novel evaluation protocol. Our approach focuses on the ability to provide early warnings of impending anomalies, rather than merely identifying them post-occurrence. 
To evaluate the predictive capability, we designate the ground-truth labels of the window immediately preceding each true anomaly as anomalous.
Also, for continuous anomalies, we only consider the first window in which the anomaly manifests. Subsequent anomalous samples within the same sequence are omitted from evaluation. 
Based on prediction results and the adjusted ground-truth labels,
we employ the point adjustment strategy and then calculate the precision $P$, recall $R$, and $F_1$ score~\cite{xu2018unsupervised,su2019robust,xu2021anomaly}.
Additionally, considering the imperfection of the point adjustment strategy, we incorporate recent metrics specifically designed for time-series anomaly detection, including Affiliation Precision/Recall (Aff-P/Aff-R)~\cite{huet2022local}, Range-AUC-ROC/PR (R-A-R/R-A-P)~\cite{paparrizos2022volume}, and Volume Under the Surface ROC/PR (V\_ROC/V\_PR).

\subsubsection{Implementation Details} 
In our experiments, we set the default observation window and target window size to 100, with a sliding window step size of 50, enabling about 50-100 minutes/hours of early warning of the future abnormal events. 
We select the anomaly score threshold for identifying anomalies according to AnomTrans~\cite{xu2021anomaly}.
The number of channels in the hidden state $d_{model}$ is 512 by default and the number of heads $h$ is 8. We use the Adam optimizer with an initial learning rate of $10^{-4}$.
For different datasets, we introduce the prediction feedback module into training at different iterations.
Specifically, this module is introduced at 30,000 iterations for SMD, 4,000 for SMAP, 15,000 for SWaT, 2,500 for MSL, and 5,000 for PSM.
All experiments were implemented in PyTorch and run on a single NVIDIA GeForce RTX 3090 GPU.

\begin{table*}[htbp]
  \centering
  \caption{Ablation study results. Bare AP only uses an anomaly prediction module without both future context and the variate-time attention. Bare Fore only uses a long-term forecasting model. w/o Att uses two separate self-attention modules. w/o $\mathcal{L}_c$ removes the future context reconstruction module.
  }
  \scalebox{0.90}{
    \begin{tabular}{l|rrr|rrr|rrr|rrr|rrr}
    \toprule
    \toprule
    \multicolumn{1}{c|}{\multirow{2}[4]{*}{Architecture}} & \multicolumn{3}{c|}{SMD} & \multicolumn{3}{c|}{MSL} & \multicolumn{3}{c|}{SMAP} & \multicolumn{3}{c|}{SWaT} & \multicolumn{3}{c}{PSM} \\
\cmidrule{2-16}          & \multicolumn{1}{c}{P} & \multicolumn{1}{c}{R} & \multicolumn{1}{c|}{$F_1$} & \multicolumn{1}{c}{P} & \multicolumn{1}{c}{R} & \multicolumn{1}{c|}{$F_1$} & \multicolumn{1}{c}{P} & \multicolumn{1}{c}{R} & \multicolumn{1}{c|}{$F_1$} & \multicolumn{1}{c}{P} & \multicolumn{1}{c}{R} & \multicolumn{1}{c|}{$F_1$} & \multicolumn{1}{c}{P} & \multicolumn{1}{c}{R} & \multicolumn{1}{c}{$F_1$} \\
    \midrule
    Bare AP & 25.19  & 34.46  & 29.10  & 39.71  & 57.86  & 47.10  & 30.51  & 51.51  & 38.32  & 24.01  & 61.03  & 34.46  & 41.55  & 36.45  & 38.84  \\
    Bare Fore & 30.57  & 45.38  & 36.53  & 45.64  & 29.69  & 35.98  & 26.19  & 42.87  & 32.51  & 6.40  & 12.50  & 8.47  & 54.28  & 28.75  & 37.59  \\
    w/o Att & 33.41  & 52.23  & 40.76  & 49.98  & 43.86  & 46.72  & 33.98  & 60.27  & 43.46  & 24.57  & \textbf{77.49} & 37.31  & 50.48  & 53.39  & 51.89  \\
    w/o $\mathcal{L}_{c}$ & 32.25  & 49.87  & 39.17  & 55.90  & 55.58  & 55.74  & 38.28  & 72.34  & 50.06  & 21.93  & 61.65  & 32.35  & 53.45  & 59.31  & 56.23  \\
    FCM   & \textbf{34.23} & \textbf{54.13} & \textbf{41.93} & \textbf{65.50} & \textbf{82.90} & \textbf{73.18} & \textbf{39.46} & \textbf{76.09} & \textbf{51.97} & \textbf{25.62} & 74.97  & \textbf{38.19} & \textbf{56.07} & \textbf{65.92} & \textbf{60.60 } \\
    \bottomrule
    \bottomrule
    \end{tabular}%
    }
  \label{tab:abla}%
\end{table*}%

\subsection{Main Results}
We conduct our anomaly prediction approach and sixteen competing models on five multivariate real-world datasets. 
Table~\ref{tab:result} shows the precision, recall, and $F_1$ score of our method FCM and its contenders. 
Please note that this experiment assesses anomaly prediction results, wherein each target window's evaluation is based on predictions from the preceding window, diverging from the conventional anomaly detection paradigm. Therefore, this task is extremely challenging, resulting in generally lower $F_1$ scores across all methods.
Nevertheless, FCM still significantly outperforms competing methods on all five datasets, demonstrating the superiority in anomaly prediction.
Notably, FCM gains remarkable recall rates, with an average $R$ exceeding 70\% across five datasets. 
Compared to the best-performing existing methods, our approach achieve 10\%, 14.5\%, 9.2\%, 7.9\%, and 21.4\% $F_1$ score improvement across the five datasets. 
In particular, on the PSM dataset, FCM illustrates significant improvement compared to the competitors, realizing an average improvement of 48.8\% in $F_1$ score. 
Among the baseline methods, BeatGAN and AnomTrans exhibit noteworthy prediction performance.
BeatGAN's strength lies in capturing high-order patterns of normal data through adversarially generated samples, while AnomTrans leverages its anomaly discrimination method to excel in criterion correlation difference. 
Nonetheless, our method still illustrates consistent superiority, mainly attributed to its ability to identify subtle yet unusual differences in preceding windows and its advantage in forecasting future anomalies.
Since the point adjustment strategy of time series anomaly detection is controversial, some studies propose new evaluation metrics, such as affiliation precision/recall and the volume under the surface (VUS), which are considered to provide a more objective assessment of detection performance.
Table~\ref{tab:vus} reports the results of new indicators, in which we focus on the comparison between FCM and two best-performing contenders under traditional metrics (see Table~\ref{tab:all metrics} in the appendix~\ref{sec:all metric} for the results of other methods). 
The empirical results suggest that FCM can also outperform these competitors across most evaluation scenarios. 

We further relax the evaluation criteria by incorporating the previously omitted anomaly segments as introduced in Section \ref{sec:metrics}. Specifically, we consider the prediction results to be accurate if they successfully trigger an alert within the extended temporal window.
It can be observed from Table~\ref{tab:detect} that our method also significantly outperforms existing state-of-the-art anomaly detectors, averagely achieving 32.3\%, 21.9\%, 24.3\%, 22.2\%, and 17.0\% improvement across five datasets, respectively. 
These empirical results underscore the efficacy of FCM in the common anomaly detection task, showcasing the strong ability of FCM in capturing both subtle and substantial abnormal signals in the respective observation and target windows. 

\subsection{Ablation Study}

This experiment aims to validate the effectiveness of several key modules in our FCM model. By removing components of FCM, we create four ablated variants, including \textbf{Bare AP}, \textbf{Bare Fore}, \textbf{w/o $\mathcal{L}_c$}, and \textbf{w/o Att}. Table~\ref{tab:abla} presents the performance of these ablated variants against the full FCM model. We have the following findings:
\begin{itemize}
    \item \textbf{Bare AP} uses a bare anomaly prediction module without future context or var-time attention. 
    We use AnomTrans to accomplish this variant. Bare AP shows a performance gap of an average of 15.6\% in $F_1$ score across five datasets, compared to the full FCM model, demonstrating the ineffectiveness of modeling the subtle normality in the observation window only.

    \item \textbf{Bare Fore} utilizes a long-term forecasting model to generate predictions for the observation window, which are then used directly for anomaly detection. Results indicate that the prediction results also contain key features for identifying future anomalies. It is worthwhile to enhance the identification of future anomalies from future context.
    

    \item \textbf{The w/o Att variant} represents the complete form of our model, yet utilizing two separate self-attention modules for the anomaly prediction and time series forecasting.
    Compared to \textbf{FCM}, this variant demonstrates an average improvement of 9.2\% in $F_1$ score.
    It can be seen that the shared self-attention mechanism effectively learns multi-dimensional normality correlation information, leading to further improved anomaly prediction performance.

    \item \textbf{The w/o $\mathcal{L}_{c}$ variant} incorporates both the time series forecasting module and the anomaly prediction module. In this configuration, our anomaly prediction module and time series forecasting module share the var-time attention mechanism. 
    This variant, on five datasets, improves the $F_1$ score by 6.5\% on average compared to \textbf{FCM}.
    This underscores the importance of future contextual information in obtaining discriminative representations of future anomalies.

\end{itemize}


In conclusion, the final row in Table~\ref{tab:abla} represents the complete form of our method, incorporating all four modules. The inclusion of both the anomaly prediction and time series forecasting modules allows the model to leverage future context and learn the normality from two complementary views, resulting the superior anomaly prediction results.

\begin{figure*}[ht]
    \centering
    \begin{subfigure}[b]{0.31\textwidth}
\includegraphics[width=\textwidth]{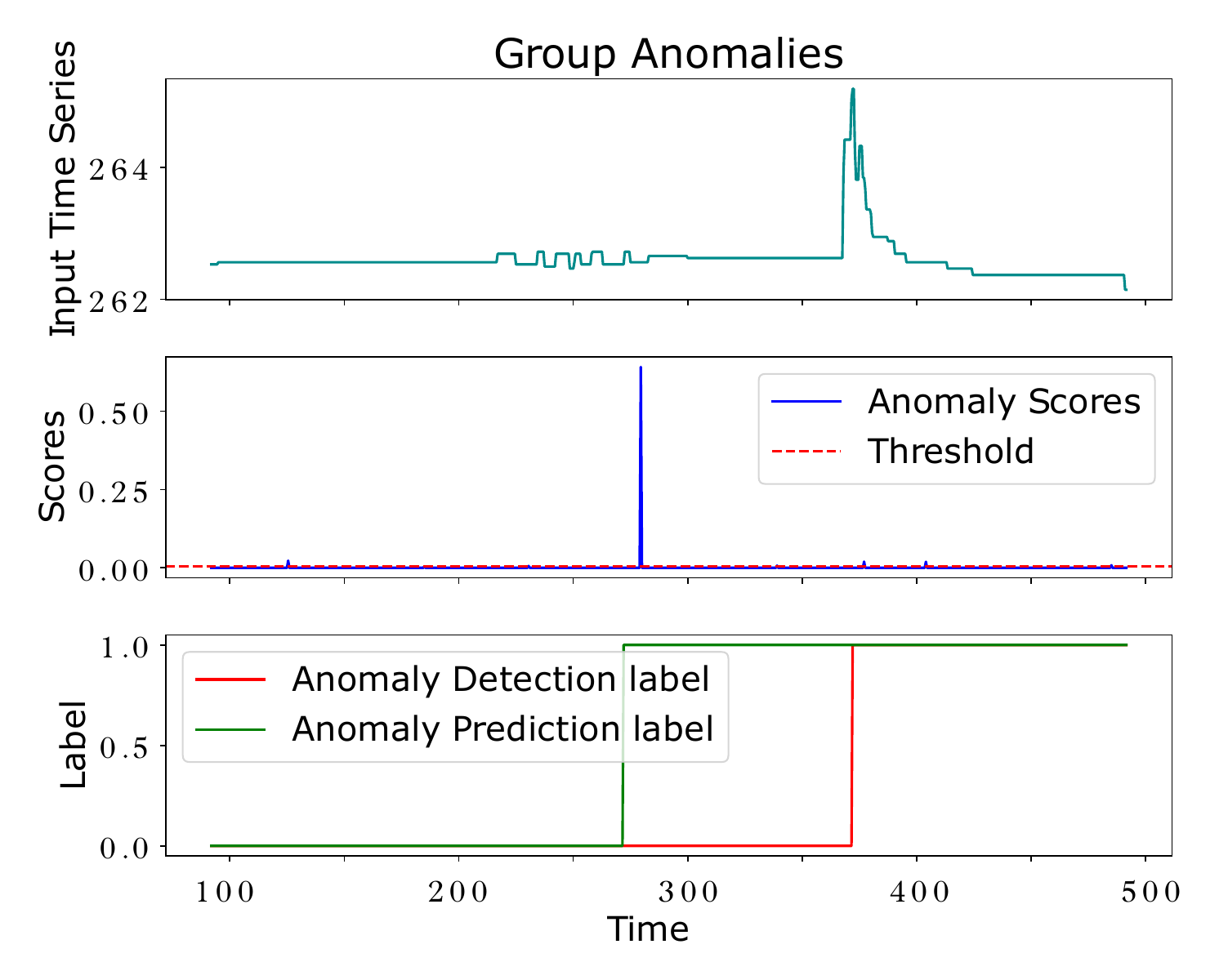}
        \caption{}
    \end{subfigure}
    \begin{subfigure}[b]{0.31\textwidth}
        \includegraphics[width=\textwidth]{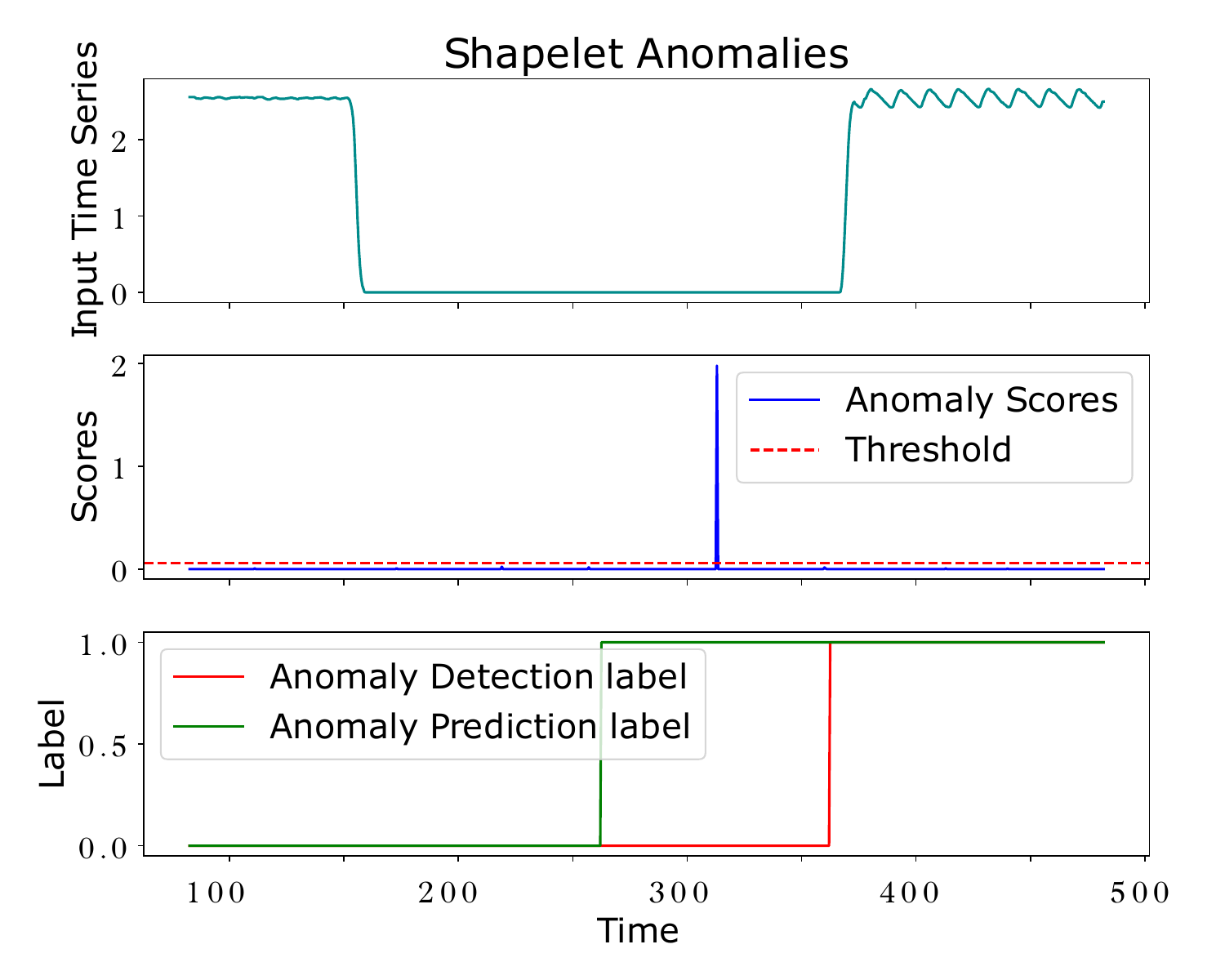}
        \caption{}
    \end{subfigure}
    \begin{subfigure}[b]{0.31\textwidth}
        \includegraphics[width=\textwidth]{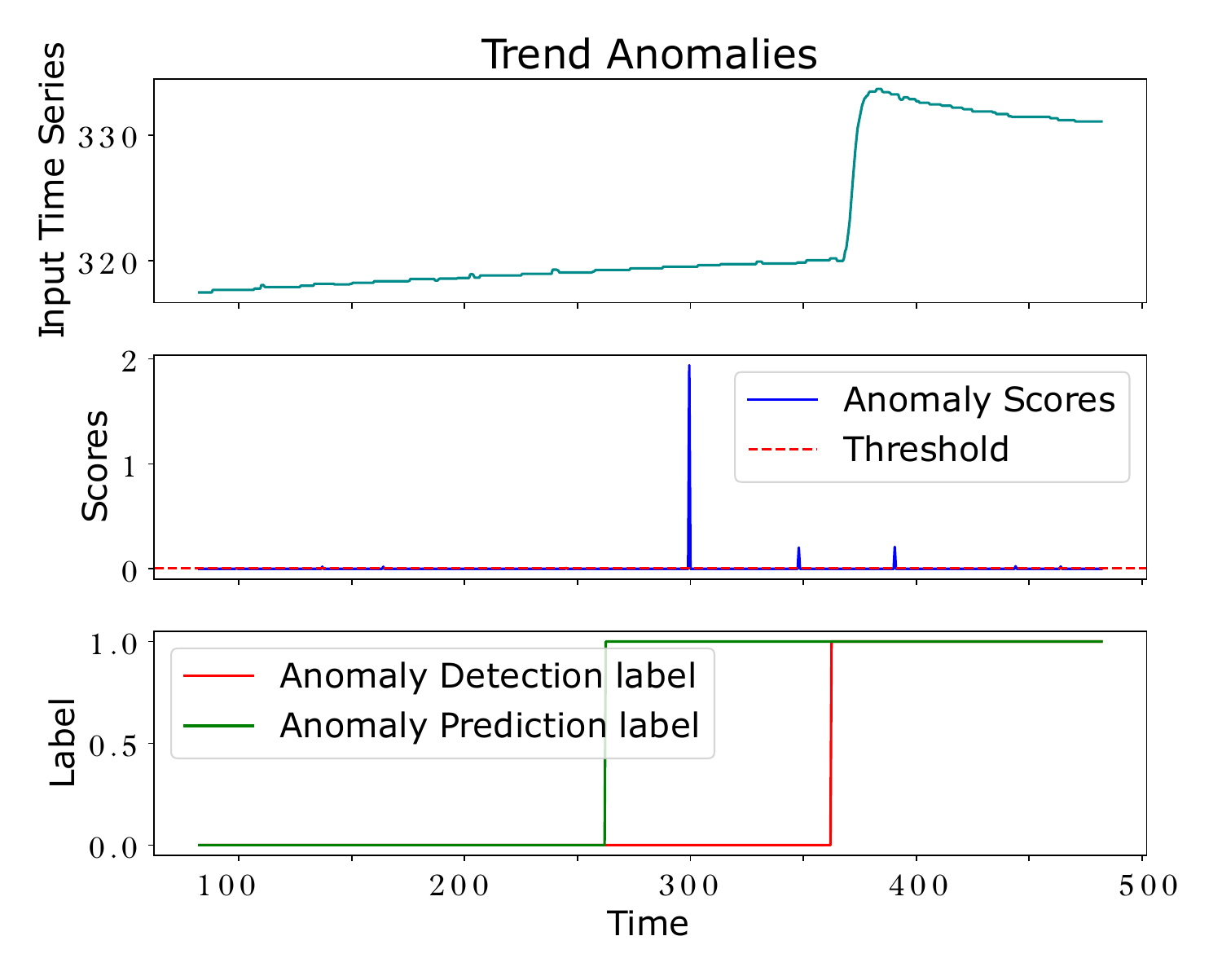}
        \caption{}
    \end{subfigure}
    \caption{Visualization of ground-truth anomalies and anomaly scores for different types of anomalies.}
\label{fig:visual}
\end{figure*}

\begin{figure*}[ht]
    \centering
    \begin{subfigure}[b]{0.31\textwidth}
\includegraphics[width=\textwidth]{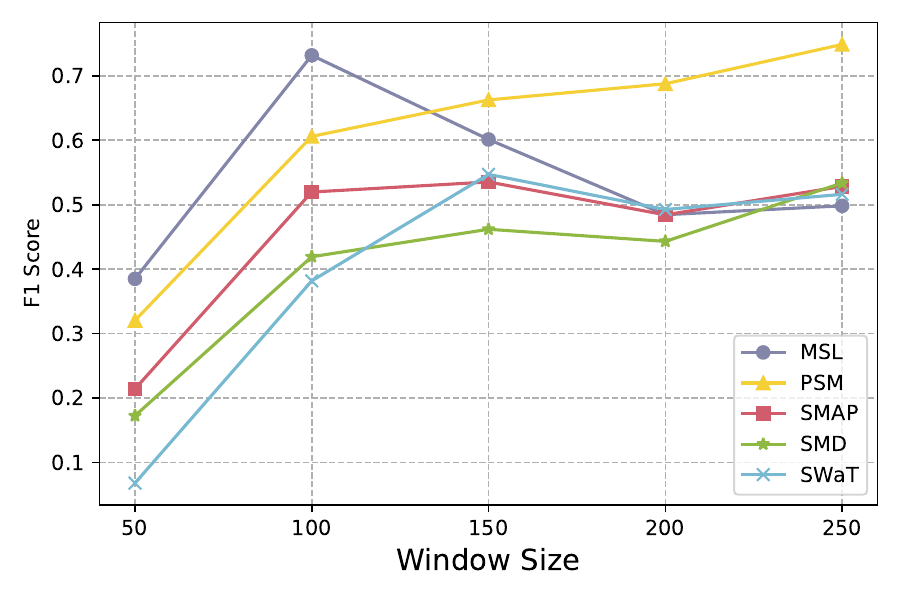}
        \caption{}
    \end{subfigure}
    \begin{subfigure}[b]{0.31\textwidth}
        \includegraphics[width=\textwidth]{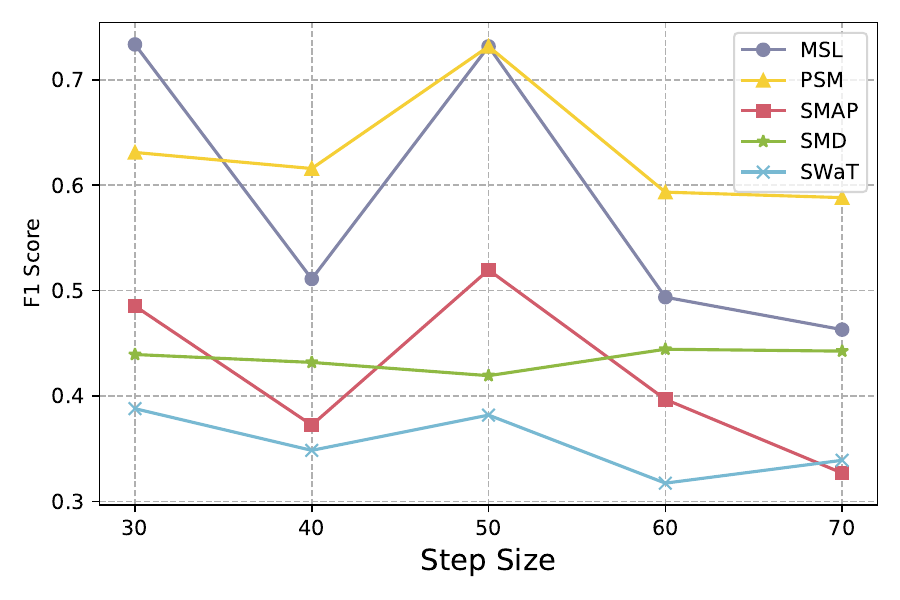}
        \caption{}
    \end{subfigure}
    \begin{subfigure}[b]{0.31\textwidth}
        \includegraphics[width=\textwidth]{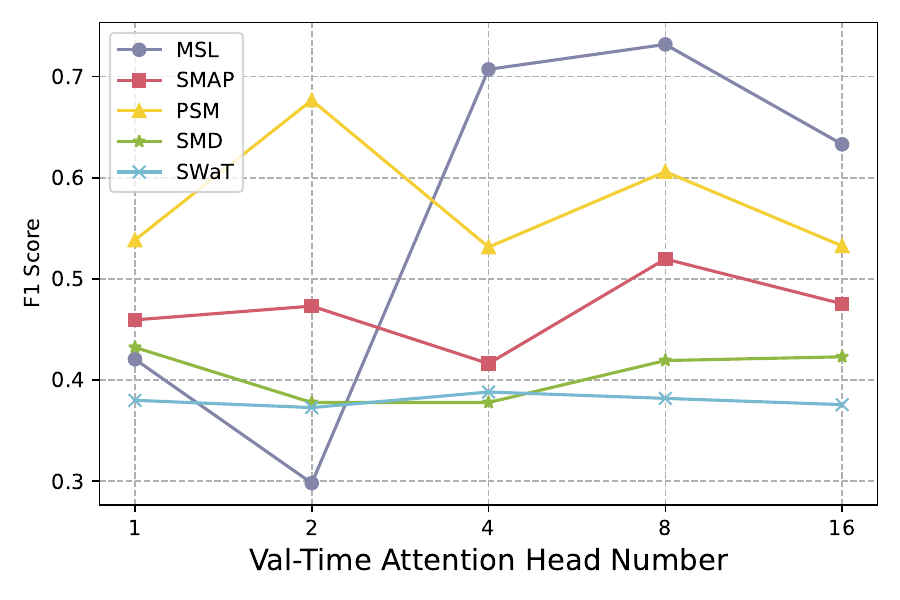}
        \caption{}
    \end{subfigure}
    \caption{Sensitivity analysis of main hyper-parameters in FCM.}
\label{fig:hype}
\end{figure*}

\subsection{Case Study}

To demonstrate the efficacy of our anomaly prediction method, we conduct a comparative case study to complement quantitative anomaly detection results. This study leverages three distinct types of anomalies commonly encountered in time series data, including group anomalies, shapelet anomalies, and trend anomalies, as respectively shown in Figure \ref{fig:visual} (a)(b)(c). These cases are sampled from the SWaT dataset. 
For each sub-figure, the upper panel displays the original time series data. The middle panel illustrates the anomaly scores (in blue) predicted by our approach FCM and the threshold (represented by a red dashed line). The bottom panel depicts the prediction labels and the original detection labels. 
FCM successfully assigns significantly high anomaly scores before the actual anomalies occur in these three typical cases, thereby offering valuable early warnings. FCM can perceive subtle abnormality signs, during which the interactions between current observational time points and its forecasting time points are effectively modeled. 
Unlike traditional methods that detect anomalies only after they have manifested, FCM ensures a proactive rather than passive response.
The ability of preemptive detection is crucial as it allows for timely intervention and mitigation of the abnormal events.



\subsection{Hyper-parameter Sensitivity}
We conduct sensitivity analysis to investigate the impact of hyper-parameters on anomaly prediction performance, i.e., this experiment varies the setting of several hyper-parameters and reports the corresponding $F_1$ scores. Specifically, we investigate three key hyper-parameters in FCM, including the window size, the sliding window step size, and the number of attention heads. 


Fig.~\ref{fig:hype} (a) illustrates the effect of varying window sizes on the prediction results. 
Note that the size of both observation window and target window is adjusted concurrently. 
Generally, a larger window is advantageous as it encapsulates more comprehensive temporal semantics, whereas increased window size correlates with higher computational complexity and more challenging forecasting process.
FCM exhibits optimal performance stability within the 100-250 samples range.

Fig.~\ref{fig:hype} (b) shows the impact of different sliding step sizes on prediction performance. 
For fixed-size windows, a smaller sliding step can extend the effective operational time, increasing the temporal resolution of the analysis.
On the contrary, a larger sliding step results in fewer learning windows, potentially increasing prediction difficulty due to reduced data overlap.
FCM demonstrates relatively stable performance on PSM, SMD, and SwAT within the 30-70 step size range, but a gradual performance degradation can be observed on the MSL and SMAP datasets. The optimal temporal granularity might vary across datasets, reflecting differences in underlying temporal dynamics and anomaly characteristics. 


Fig.~\ref{fig:hype} (c) elucidates the impact of the number of var-time attention heads on the model performance. 
The MSL dataset is sensitive to variations in the number of attention heads, while other datasets demonstrate relative stability across different configurations.
In alignment with standard Transformer architectures, an increased number of attention heads generally brings enhanced information capture capacity. 
Following typical setting, we also determine that eight attention heads represent an optimal configuration.


\section{Conclusion}

This paper introduces a novel method for time series anomaly prediction, FCM. 
It models future context for accurate anomaly prediction by learning to capture subtle abnormality signs in the observational time points and their correlations with forecasting time points.
FCM achieves this by modeling the correlation between the current and future windows through a time series forecasting module, in which the forecasting future time points act as the feature amplifier of the current window. This helps largely enhance the expressiveness of the abnormal features. It also introduces 
a joint variate-time attention module to simultaneously learn the correlations between time series and between the features. 
Under the constraints of the two correlations, future anomalies become more difficult to reconstruct, resulting in easier differentiation between future normal and abnormal time points.
The effectiveness of FCM is justified by extensive experiments on five real-world datasets compared to 16 state-of-the-art competing methods.


\bibliographystyle{ACM-Reference-Format}
\bibliography{main}

\appendix

\section{datasets}
Five publicly available multivariate time series datasets
are used in our experiments, with the relevant statistics shown in Table~\ref{tab:dataset}.
AR (anomaly ratio) represents the abnormal proportion of the whole dataset.

\begin{table}[htbp]
  \centering
  \caption{Dataset Description}
    \begin{tabular}{l|cccc}
    \toprule
    \toprule
    Datasets & Dimension & Training & Test  & AR (\%) \\
    \midrule
    SMD   & 38    & 708,405 & 708,420 & 4.2 \\
    MSL   & 55    & 58,317 & 73,729 & 10.5 \\
    SMAP  & 25    & 135183 & 427,617 & 12.8 \\
    SWaT  & 51    & 495,000 & 449,919 & 12.1 \\
    PSM   & 25    & 132,481 & 87,841 & 27.8 \\
    \bottomrule
    \bottomrule
    \end{tabular}%
  \label{tab:dataset}%
\end{table}%

\section{ALGORITHM}
We show the specific implementation of FCM in Alg.~\ref{alg:FCM}.
Through the results of time series embedding, our attention can learn the association between variables and output the predicted results (Line~\ref{alg:forecast}).
Through the embedding of variables at the same time point, our attention can learn the association between time points and output the reconstructed results, which are used for anomaly prediction (Line~\ref{alg:ap}).
We control the training epoch in which we involve the reconstruction module of future context through a parameter $\mathcal{P}$ in Step 8. 

\begin{algorithm}[b]
\renewcommand{\algorithmicrequire}{\textbf{Input}}
\renewcommand{\algorithmicensure}{\textbf{Output}}
\caption{Learning algorithm of FCM}\label{alg:FCM}
\begin{algorithmic}[1]
\Require (1) Training data $\mathcal{X}$;
(2) Hyper-parameters: future context access point $\mathcal{P}$;
(3) Training epochs $E$ and the number of batches in every epoch $Q$.
\Ensure $\Theta_{det}$: the parameter of anomaly prediction network.

\State Initialize the LTTSF network $\Theta_{fore}$ and the anomaly prediction network $\Theta_{det}$;
\While{$e<E$}
    \For{$q<Q$}
        \State {Embed every time series of each sensor as a token;}

        \State {Put it into the encoder of transformer and output the LTTSF results;}\label{alg:forecast}
        
        \State {Embed the value of every sensor at the same time point as a token;}

        \State {Put into the encoder of transformer and output the reconstruction results;}\label{alg:ap}

        \If{$e > \mathcal{P}$}\label{alg:fc}
            \State Get the forecasting results through $\Theta_{fore}$;
            \State Update $\Theta_{det}$, $\Theta_{fore}$  with $\mathcal{L}_{f}$, $\mathcal{L}_{det}$ and $\mathcal{L}_{c}$, according to Eq.~\eqref{eq:foreloss}, Eq.~\eqref{eq:adloss} and Eq.~\eqref{eq:interloss}.
        \EndIf
        \State Update $\Theta_{det}$, $\Theta_{fore}$ with $\mathcal{L}_{f}$, $\mathcal{L}_{det}$ according to Eq.~\eqref{eq:foreloss} and Eq.~\eqref{eq:adloss}.
            
    \EndFor
\EndWhile
\State \textbf{Evaluation:}
\State Compute anomaly prediction scores according to Eq.~\eqref{eq:score}.
\end{algorithmic}
\end{algorithm}

\section{Results on different metrics}
\label{sec:all metric}
Results of anomaly prediction on 3 traditional metrics and 6 new proposed metrics are shown as Table~\ref{tab:all metrics}. 
It can be seen that our FCM achieves excellent results in most indicators, especially in PR-related indicators.
This demonstrates that FCM can effectively leverage the future context to predict future anomalies accurately.
\begin{table*}[htbp]
  \centering
  \caption{Precision $P$, Recall $R$, F1 score, Aff-P, Aff-R, R\_A\_R, R\_A\_P, V\_ROC and V\_PR of anomaly prediction results on five multivariate real-world datasets. All results are represented in percentages (\%). The best performance for each metric on each dataset is highlighted in bold.}
  \resizebox{120mm}{115mm}{

    \begin{tabular}{c|l|rrrrrrrrrr}
    \toprule
    \toprule
    \multicolumn{1}{l|}{Dataset} & Method  & \multicolumn{1}{l}{Accuracy} & \multicolumn{1}{l}{Precision} & \multicolumn{1}{l}{Recall} & \multicolumn{1}{l}{F1\_score} & \multicolumn{1}{l}{Aff-P} & \multicolumn{1}{l}{Aff-R} & \multicolumn{1}{l}{R\_A\_R} & \multicolumn{1}{l}{R\_A\_P} & \multicolumn{1}{l}{V\_ROC} & \multicolumn{1}{l}{V\_PR} \\
    \midrule
    \multirow{17}[2]{*}{SMD} & OCSVM & 75.78  & 7.06  & 11.23  & 8.67  & 51.77  & 93.62  & 63.55  & 29.66  & 63.21  & 29.28  \\
          & iForest & 75.64  & 4.52  & 3.27  & 3.80  & 50.18  & 84.04  & 57.66  & 23.68  & 56.95  & 22.92  \\
          & BeatGAN & 87.31  & 27.36  & 38.35  & 31.94  & 50.06  & \textbf{97.56} & \textbf{64.28} & 25.88  & \textbf{63.67} & 25.29  \\
          & DAGMM & 75.37  & 2.12  & 43.10  & 4.05  & 50.42  & 76.28  & 51.27  & 17.18  & 51.17  & 17.04  \\
          & OmniAnomaly & 74.08  & 0.08  & 0.17  & 0.11  & 49.58  & 68.00  & 46.15  & 11.89  & 45.93  & 11.65  \\
          & THOC  & 75.61  & 2.49  & 6.06  & 3.53  & 51.70  & 83.04  & 54.15  & 19.94  & 53.69  & 19.45  \\
          & InterFusion & 73.77  & 1.58  & 3.43  & 2.16  & 50.99  & 68.39  & 48.67  & 14.84  & 48.24  & 14.35  \\
          & LSTM-ED & 76.52  & 6.72  & 11.05  & 8.35  & 52.65  & 95.09  & 63.54  & 29.25  & 63.15  & 28.84  \\
          & MSCRED & 75.31  & 1.53  & 17.03  & 2.81  & 55.35  & 57.16  & 47.82  & 13.68  & 47.47  & 13.28  \\
          & GDN   & 75.58  & 4.31  & 8.16  & 5.64  & 55.63  & 94.27  & 60.66  & 26.67  & 60.60  & 26.58  \\
          & COUTA & 80.62  & 3.62  & 35.93  & 6.58  & 51.39  & 91.72  & 57.69  & 21.27  & 57.85  & 21.42  \\
          & FEDformer & 75.68  & 3.08  & 2.84  & 2.96  & 51.42  & 80.55  & 57.88  & 23.18  & 57.57  & 22.80  \\
          & Autoformer & 75.65  & 3.08  & 2.84  & 2.96  & 51.02  & 80.48  & 57.34  & 22.62  & 57.08  & 22.28  \\
          & TimesNet & 98.46  & 14.27  & 12.51  & 13.33  & \textbf{57.14} & 33.10  & 51.00  & 11.10  & 50.96  & 11.12  \\
          & AnomTrans & 98.41  & 25.19  & 34.46  & 29.10  & 49.39  & 90.09  & 57.04  & 22.07  & 56.12  & 21.15  \\
          & DCdetector & 98.25  & 17.14  & 21.53  & 19.08  & 49.79  & 90.12  & 55.17  & 16.55  & 54.40  & 15.78  \\
          & FCM (Ours)  & \textbf{98.58} & \textbf{34.23} & \textbf{54.13} & \textbf{41.93} & 50.46  & 93.24  & 61.37  & \textbf{30.79} & 59.96  & \textbf{29.33 } \\
    \midrule
    \multirow{17}[2]{*}{MSL} & OCSVM & 76.37  & 39.21  & 46.15  & 42.40  & 50.59  & 99.14  & 75.89  & 44.78  & 76.92  & 45.79  \\
          & iForest & 76.74  & 38.94  & 20.34  & 26.72  & 50.57  & 99.09  & 75.41  & 44.14  & 76.56  & 45.28  \\
          & BeatGAN & 82.90  & 58.36  & 58.99  & 58.67  & 49.81  & 98.56  & 72.44  & 38.90  & 73.69  & 40.22  \\
          & DAGMM & 76.12  & 47.30  & 53.24  & 50.10  & 50.57  & 99.15  & 75.99  & 44.96  & 76.89  & 45.83  \\
          & OmniAnomaly & 74.83  & 10.09  & 10.07  & 10.08  & 49.98  & 95.83  & 57.93  & 25.57  & 56.74  & 24.42  \\
          & THOC  & 76.24  & 23.55  & 35.27  & 28.24  & 50.26  & 98.63  & 75.66  & 44.69  & 76.50  & 45.49  \\
          & InterFusion & 74.85  & 23.65  & 28.38  & 25.80  & 49.40  & 81.94  & 58.69  & 27.02  & 58.81  & 27.14  \\
          & LSTM-ED & 75.09  & 37.12  & 42.36  & 39.57  & 48.73  & 98.89  & 73.10  & 42.28  & 73.83  & 42.97  \\
          & MSCRED & 78.20  & 6.26  & 24.05  & 9.94  & 52.45  & 90.89  & 60.03  & 27.31  & 60.56  & 27.81  \\
          & GDN   & 75.80  & 24.85  & 38.51  & 30.21  & 51.65  & 98.61  & 73.79  & 42.44  & 72.99  & 41.66  \\
          & COUTA & 81.15  & 36.37  & 35.61  & 35.98  & 50.69  & 98.67  & 74.56  & 41.98  & 74.94  & 42.38  \\
          & FEDformer & 76.07  & 43.52  & 24.22  & 31.12  & 50.88  & 97.37  & 76.03  & 44.04  & 75.47  & 43.45  \\
          & Autoformer & 75.98  & 43.01  & 23.68  & 30.55  & 51.02  & \textbf{99.26} & \textbf{81.11} & 49.41  & \textbf{80.86} & 49.11  \\
          & TimesNet & 97.44  & 39.00  & 24.22  & 29.89  & \textbf{53.44} & 77.29  & 54.50  & 28.48  & 54.55  & 28.32  \\
          & AnomTrans & 98.04  & 39.71  & 57.86  & 47.10  & 50.34  & 93.84  & 66.27  & 46.88  & 65.09  & 45.76  \\
          & DCdetector & 97.04  & 21.17  & 11.50  & 14.91  & 47.38  & 83.73  & 53.11  & 17.46  & 52.89  & 17.34  \\
          & FCM (Ours)  & \textbf{98.66} & \textbf{65.50} & \textbf{82.90} & \textbf{73.18} & 50.51  & 96.12  & 75.34  & \textbf{59.81} & 74.63  & \textbf{59.14 } \\
    \midrule
    \multirow{17}[2]{*}{SMAP} & OCSVM & 94.57  & 23.41  & 34.16  & 27.78  & 42.21  & 74.98  & 58.93  & 17.70  & 59.56  & 18.33  \\
          & iForest & 76.09  & 29.28  & 46.29  & 35.87  & 45.35  & 86.28  & 61.50  & 25.98  & 61.52  & 25.98  \\
          & BeatGAN & 92.17  & 34.49  & 56.37  & 42.80  & 50.09  & \textbf{99.43} & \textbf{74.17} & 33.81  & \textbf{74.86} & 34.52  \\
          & DAGMM & 74.80  & 20.78  & 59.13  & 30.75  & 42.76  & 83.61  & 59.21  & 24.19  & 59.93  & 24.92  \\
          & OmniAnomaly & 74.91  & 2.55  & 5.83  & 3.55  & 49.93  & 98.32  & 59.81  & 24.46  & 59.35  & 23.98  \\
          & THOC  & 73.65  & 21.39  & 30.56  & 25.17  & 47.25  & 87.52  & 59.39  & 24.89  & 60.40  & 25.90  \\
          & InterFusion & 75.89  & 19.71  & 54.09  & 28.89  & 49.47  & 90.67  & 66.75  & 31.32  & 66.60  & 31.16  \\
          & LSTM-ED & 77.19  & 14.37  & 39.12  & 21.02  & 45.59  & 90.41  & 61.76  & 25.69  & 62.33  & 26.25  \\
          & MSCRED & 75.20  & 3.65  & 41.46  & 6.71  & 46.00  & 85.41  & 57.65  & 22.36  & 57.45  & 22.15  \\
          & GDN   & 74.03  & 17.06  & 45.53  & 24.82  & 44.23  & 84.94  & 58.24  & 23.51  & 59.16  & 24.42  \\
          & COUTA & 79.19  & 14.56  & 37.86  & 21.03  & 44.00  & 84.17  & 59.88  & 22.91  & 60.56  & 23.57  \\
          & FEDformer & 74.25  & 15.87  & 21.25  & 18.17  & 50.39  & 91.74  & 65.91  & 30.65  & 65.54  & 30.27  \\
          & Autoformer & 74.17  & 14.02  & 19.13  & 16.18  & 49.62  & 88.03  & 65.20  & 30.11  & 64.97  & 29.87  \\
          & TimesNet & 98.52  & 19.76  & 24.24  & 21.77  & 48.91  & 61.48  & 53.53  & 16.82  & 53.33  & 16.53  \\
          & AnomTrans & 98.50  & 30.51  & 51.51  & 38.32  & \textbf{50.54} & 97.97  & 66.40  & 33.02  & 65.70  & 32.32  \\
          & DCdetector & 98.20  & 6.82  & 8.44  & 7.54  & 49.31  & 95.10  & 52.98  & 8.30  & 52.95  & 8.33  \\
          & FCM (Ours)  & \textbf{98.81} & \textbf{39.46} & \textbf{76.09} & \textbf{51.97} & 49.95  & 97.95  & 73.97  & \textbf{44.77} & 72.63  & \textbf{43.45 } \\
    \midrule
    \multirow{17}[2]{*}{SWaT} & OCSVM & 80.76  & 6.72  & 17.92  & 9.77  & 52.09  & 46.04  & 44.61  & 5.04  & 44.42  & 4.85  \\
          & iForest & 81.12  & 7.34  & 10.19  & 8.53  & 52.72  & 84.42  & 49.45  & 9.94  & 50.17  & 10.64  \\
          & BeatGAN & 82.84  & 15.05  & 39.67  & 21.82  & 49.77  & \textbf{99.23} & 73.60  & 33.69  & \textbf{74.06} & 34.15  \\
          & DAGMM & 80.72  & 0.26  & 7.78  & 0.50  & 56.12  & 50.96  & 41.53  & 1.64  & 41.59  & 1.70  \\
          & OmniAnomaly & 71.95  & 0.37  & 14.02  & 0.72  & 43.67  & 67.21  & 41.33  & 5.98  & 40.99  & 5.63  \\
          & THOC  & 81.72  & 0.61  & 24.79  & 1.18  & 52.76  & 86.39  & 46.86  & 6.80  & 47.00  & 6.92  \\
          & InterFusion & 72.35  & 0.60  & 15.51  & 1.16  & 49.30  & 56.31  & 50.39  & 15.22  & 50.27  & 15.10  \\
          & LSTM-ED & 80.73  & 0.83  & 25.36  & 1.60  & 55.01  & 46.88  & 46.59  & 7.15  & 46.43  & 6.98  \\
          & MSCRED & 81.62  & 0.56  & 6.15  & 1.02  & 42.42  & 15.07  & 42.05  & 1.80  & 41.91  & 1.66  \\
          & GDN   & 82.60  & 3.08  & 21.64  & 21.64  & 50.82  & 90.76  & 56.77  & 16.71  & 55.89  & 15.83  \\
          & COUTA & 78.06  & 0.14  & 3.38  & 0.27  & 44.95  & 35.33  & 42.39  & 4.05  & 42.33  & 3.99  \\
          & FEDformer & 80.87  & 4.08  & 21.16  & 6.85  & 51.42  & 56.93  & 45.27  & 5.53  & 45.14  & 5.39  \\
          & Autoformer & 80.87  & 4.07  & 21.16  & 6.82  & 48.35  & 56.26  & 45.71  & 5.96  & 45.39  & 5.64  \\
          & TimesNet & 98.88  & 12.97  & 26.07  & 17.32  & \textbf{56.39} & 63.00  & 53.39  & 12.25  & 53.62  & 12.48  \\
          & AnomTrans & \textbf{98.92} & 24.01  & 61.03  & 34.46  & 50.27  & 87.15  & 71.14  & 34.85  & 71.66  & 35.39  \\
          & DCdetector & 98.62  & 5.49  & 13.26  & 7.77  & 50.67  & 95.60  & 54.58  & 8.53  & 54.68  & 8.67  \\
          & FCM (Ours)  & 98.91  & \textbf{25.62} & \textbf{74.97} & \textbf{38.19} & 50.64  & 95.35  & \textbf{74.11} & \textbf{37.71} & 73.03  & \textbf{36.64 } \\
    \midrule
    \multirow{17}[2]{*}{PSM} & OCSVM & 80.71  & 3.51  & 26.30  & 6.19  & 53.18  & 62.94  & 49.78  & 15.88  & 50.03  & 16.05  \\
          & iForest & 80.39  & 17.34  & 23.17  & 19.84  & 48.73  & 82.25  & 59.66  & 27.11  & 59.75  & 27.05  \\
          & BeatGAN & 91.82  & 42.30  & 36.54  & 39.21  & 50.16  & \textbf{93.90} & \textbf{66.97} & 33.46  & \textbf{65.97} & 32.36  \\
          & DAGMM & 78.94  & 5.66  & 45.18  & 10.06  & 48.88  & 84.37  & 60.15  & 27.72  & 59.18  & 26.75  \\
          & OmniAnomaly & 75.44  & 15.77  & 13.00  & 14.25  & 51.90  & 80.40  & 53.84  & 22.20  & 54.21  & 22.49  \\
          & THOC  & 78.48  & 2.60  & 12.23  & 4.29  & \textbf{60.87} & 39.41  & 47.84  & 14.41  & 47.51  & 14.02  \\
          & InterFusion & 66.36  & 4.85  & 3.36  & 3.97  & 45.96  & 69.45  & 51.88  & 24.14  & 52.06  & 24.28  \\
          & LSTM-ED & 81.82  & 5.07  & 12.00  & 7.13  & 56.79  & 61.06  & 49.11  & 14.71  & 49.66  & 15.14  \\
          & MSCRED & 77.91  & 2.85  & 22.02  & 5.05  & 51.68  & 40.00  & 46.74  & 13.42  & 46.26  & 12.90  \\
          & GDN   & 79.63  & 1.97  & 6.35  & 3.01  & 58.98  & 62.87  & 51.16  & 17.80  & 51.59  & 18.12  \\
          & COUTA & 82.73  & 2.49  & 19.34  & 4.41  & 59.24  & 14.42  & 45.92  & 10.00  & 45.69  & 9.71  \\
          & FEDformer & 79.03  & 5.24  & 12.83  & 7.44  & 56.26  & 45.16  & 47.16  & 12.03  & 46.98  & 11.74  \\
          & Autoformer & 79.16  & 5.18  & 12.83  & 7.38  & 57.83  & 37.04  & 46.32  & 11.10  & 46.06  & 10.74  \\
          & TimesNet & 97.87  & 30.21  & 10.80  & 15.91  & 56.83  & 41.86  & 51.76  & 20.90  & 51.69  & 20.99  \\
          & AnomTrans & 97.87  & 41.55  & 36.45  & 38.84  & 51.54  & 78.03  & 58.54  & 32.71  & 57.70  & 31.94  \\
          & DCdetector & 98.03  & 7.91  & 8.02  & 7.96  & 50.45  & 84.80  & 52.61  & 11.75  & 52.31  & 11.18  \\
          & FCM (Ours)  & \textbf{98.40} & \textbf{56.07} & \textbf{65.92} & \textbf{60.60} & 49.86  & 80.15  & 65.93  & \textbf{46.82} & 64.24  & \textbf{45.06 } \\
    \bottomrule
    \end{tabular}%

    }
  \label{tab:all metrics}%
\end{table*}%

\end{document}